\pdfoutput=1
\documentclass[11pt]{article}

\usepackage{ACL2023}

\usepackage{times}
\usepackage{latexsym}

\usepackage[T1]{fontenc}
\usepackage[utf8]{inputenc}
\usepackage{booktabs} % For formal tables
\usepackage{tabularx}
\usepackage{amsmath}
\usepackage{amssymb}
\usepackage{mathrsfs}
\usepackage{graphicx}
\usepackage{subcaption}
\usepackage{float}
\usepackage{multirow}
\usepackage{seqsplit}
\usepackage{longtable}

\usepackage{microtype}

\usepackage{inconsolata}

\usepackage{amsmath}
\title{{Regex-augmented Domain Transfer Topic Classification based on a Pre-trained Language Model: An application in Financial Domain}}

\author{Vanessa Liao \\
    Manulife \\
  250 Bloor St E \\
  Toronto, ON M4W 1E5, Canada \\
  \texttt{Vanessa\_Liao@manulife.com} \\\And
  Syed Shariyar Murtaza \\
    Manulife \\
  250 Bloor St E \\
  Toronto, ON M4W 1E5, Canada \\
  \texttt{Syed\_Shariyar\_Murtaza@manulife.com} 
  \\\AND
  Yifan Nie \\
    Manulife \\
  250 Bloor St E \\
  Toronto, ON M4W 1E5, Canada \\
  \texttt{Yifan\_Nie@manulife.com} \\\And
  Jimmy Lin \\
  University of Waterloo \\
  200 University Avenue West \\
  Waterloo, ON N2L 3G1 Canada \\
  \texttt{jimmylin@uwaterloo.ca} \\}

\begin{document}
{\makeatletter\acl@finalcopytrue
  \maketitle
}
\begin{abstract}
%A common way to use large pre-trained language models for downstream tasks is to fine tune them using additional layers. This may not work well if downstream domain is a specialized domain and a large language model has been pre-trained on a general corpus. In this paper, we discuss the use of regular expressions based patterns as features for domain knowledge during the time of fine tuning, in addition to domain specific text.  We show that this method of fine tuning improves the downstream text classification tasks as compared to fine tuning only on domain specific text. We also show that the use of attention network for fine tuning improves results compared to simple linear layers.
A common way to use large pre-trained language models for downstream tasks is to fine tune them using additional layers. This may not work well if downstream domain is a specialized domain whereas the large language model has been pre-trained on a generic corpus. In this paper, we discuss the use of regular expression patterns employed as features for domain knowledge during the process of fine tuning, in addition to domain specific text. Our experiments on real scenario production data show that this method of fine tuning improves the downstream text classification tasks as compared to fine tuning only on domain specific text. We also show that the use of attention network for fine tuning improves results compared to simple linear layers.
\end{abstract}

\section{Introduction}

In industry, a very popular application of textual data analysis is topic labeling or topic classification. Textual data could be in the form of surveys, messages on online forums or voice call transcripts. Assigning topics (or themes) to this textual data helps business decision makers understand common issues or themes being discussed by users of company's products. A common method to solve this problem is the use of topic modeling algorithms \cite{DBLP:journals/csur/ChauhanS22, DBLP:journals/corr/abs-2203-05794}. These methods generate cluster of words from which a topic needs to be inferred; e.g., concatenation of top k words. Sometimes top words may not be a good representation and human judgment is needed. In other occasions, top words may overlap with other clusters. This is often not very helpful in the business context if business users are looking for specific topics (themes) and not dynamic topics (themes); e.g.,  specific insurance claim results, complains on call center agents' attitude, lack of follow up on customer complains, etc. Another way to solve the problem of understanding common topics in large amount of data is through text classification. However, this requires labeling of textual data for many different topics. Usually manually labeling large amount of data is not feasible. A pre-trained model can be fine-tuned on a smaller amount of labelled data and then can be used to assign topics to textual artifacts \cite{DBLP:conf/cncl/SunQXH19}. When a pre-trained language model is fine tuned on small dataset, usually an additional layer is added on top of the pre-trained model and parameters of this layer are updated using a standard training method of neural classifier \cite{DBLP:conf/cncl/SunQXH19}. Although pre-trained models have demonstrated their huge potentials in many academic test benchmarks \cite{DBLP:journals/corr/abs-2211-08073}, we note that in specialized domain, simple fine tuning of pre-trained model on smaller domain-special datasets yields below par results. This is possibly due to the fact that most pre-trained language models are trained on generic corpus like Wikipedia and books \cite{DBLP:conf/naacl/DevlinCLT19, DBLP:journals/corr/abs-1907-11692}, and thus have seen few highly specialized terminology which frequently occurs in industry applications. 

\begin{figure}[htbp]
  \centering
  \includegraphics[width=0.45\textwidth]{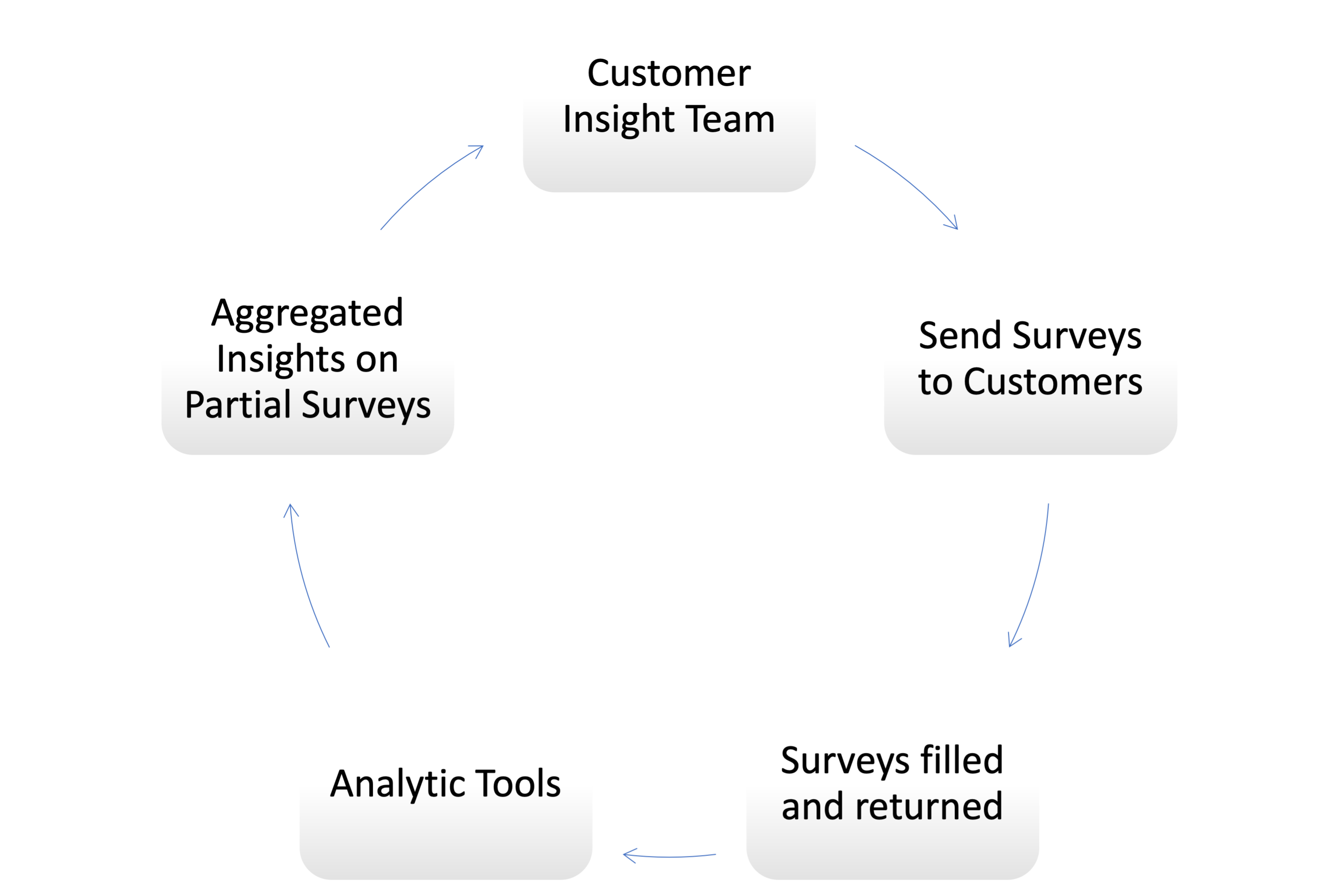}
  \caption{Survey Analytic Modeling Process}
  \label{fig:process}
\end{figure}

In this paper, we focus on fine tuning a large pre-trained language model on domain specific text, with limited amount of labelled data and large number of topics, in the context of a very large multinational insurance company and a financial institution. In this organization, a customer insight team sends out surveys to customers who have contacted the company's call centres in the past for different issues. When the surveys are returned, customer insight team runs predefined key word searches (or pattern based searches) using automated tools to generate aggregated insights on surveys, such as identifying most common issues of customers. The process is depicted in Fig \ref{fig:process}. These insights miss many potential customer issues because keyword searches have a problem of vocabulary mismatch and unseen patterns of phrases are not captured. 

Specifically, we discuss a method to fine-tune a language model on a smaller insurance and financial domain specific dataset, assisted by a set of regular expression based rules as auxiliary input.  We first tag textual data (surveys) using 27 different regular expressions (one regular expression per topic). Second we convert these tags into embeddings and fine-tune the model using both original text of surveys and regular expression based embeddings. Third, we enter the original text into a pre-trained BERT encoder \footnote{https://huggingface.co/bert-base-cased}, and fuse the encoded text representations with regex-based embeddings. Finally we fine-tune this model on 379 human labelled surveys and our results show that it outperforms simple method of fine-tuning by 20.75\% in F-1 score (see Table \ref{tab:results}) on real scenario production dataset in specialized domains. Besides improvements on those objective metrics, we also conducted user satisfaction study (described in Section \ref{sec:eval}), which shows significant improvements compared against the old system and confirms the effectiveness of our model when deployed into a production application and evaluated by end users.

%Specifically, we discuss a method to fine tune a language model on a smaller insurance and financial domain specific textual dataset using a set of regular expressions based rules as input embeddings. Our method uses a a pre-trained BERT-base-cased \footnote{https://huggingface.co/bert-base-cased} model trained on BooksCorpus and English Wikipedia corpus \cite{DBLP:conf/naacl/DevlinCLT19}. We first tag textual data (surveys) using 27 different regular expressions (one regular expression per topic). Second we convert these tags into embeddings and fine tune the model using both original text of surveys and regular expressions based embeddings. We fine tune this model on 379 human labelled surveys and our results show that it outperforms simple method of fine tuning by 20.75\% in F-1 score (see Table \ref{tab:results}) on specialized domains.

The main contribution of this paper is the demonstration of the use of a pre-trained language model in financial domain using regular expression-based embeddings alongside textual embeddings. This method is specially useful when limited label dataset is available and which is the most common problem in industry. This paper also demonstrates that the use of attention network rather than linear layers is a better option. The rest of the paper discusses related work, methodology, experimental setup, evaluation and conclusion.

\section{Related Work} \label{sec:related_work}

Topic classification, which aims to predict the topic label from a given text, is a very important task in many NLP applications. Usually, topic classification models could be roughly categorized into two types: non-neural models and neural models. Non-neural models usually employ traditional classifiers such as Naive-Bayes and SVM \cite{DBLP:conf/acl/WangM12, DBLP:journals/ml/RubinCSS12}. Those non-neural models often rely on sparse features based on term frequencies over the whole vocabulary space, which lacks the power to generalize in semantic space. To address this issue, neural models attempt to learn dense semantic representations from raw texts  and then apply a softmax classification layer to predict the topic label. For example Albatayha et al. \cite{albatayha2021multi} employs LSTM to learn the semantic representation of the texts. In \cite{DBLP:conf/www/PengLHLBWS018}, CNN is employed as the representation learning component to build semantic representations.

Recently, the rapid development of pre-trained language models \cite{DBLP:conf/naacl/DevlinCLT19, DBLP:conf/nips/YangDYCSL19, DBLP:conf/acl/LewisLGGMLSZ20} have demonstrated their potentials in many NLP tasks. The advantage of pre-trained language model lies in the fact that the costly training of representations could be done off-line in advance, and the end user only needs to fine-tune the pre-trained language model together with the task-specific layers with much smaller datasets for the downstream tasks. This idea is not new, and have been adopted in a few studies: Ghourabi et al. \cite{ghourabi2021bert} utilizes BERT \cite{DBLP:conf/naacl/DevlinCLT19} as the pre-trained language model and applies a dense layer to classify topics. Liu et al. \cite{liu2020bert} combines BERT and LSTM to conduct topic modeling task.

Although those approaches based on pre-trained language models have demonstrated their potentials, they were mostly applied and evaluated on general domains such as social media, news articles and product reviews. In fact, one of the biggest challenges for applying pre-trained language model on topic classification is domain adaptation. Given a highly specialized domain (finance and insurance in our case), and very little labeled data, it would be hard to fine-tune the model effectively. Previous work tries to achieve domain adaptation either by designing auxiliary fine-tuning structures on top of the pre-trained language model \cite{DBLP:conf/complexnetworks/MozafariFC19} or by re-training the pre-trained language model on domain-specific corpus and then fine-tuning it with classification layer \cite{DBLP:conf/interspeech/WhangLLYOL20}. 

We argue that for highly specialized domains, a more effective domain adaptation strategy is to exploit as much as possible the domain knowledge. In fact, in those domain specific applications, there are often certain rules or patterns which could help the model to detect the topic. One of the strategies to extract and match those patterns from raw texts is to apply regular expression. However, we observe that in the literature, there is still a gap between the direct application of pattern extractors such as regular expression and fine-tuning pre-trained language models for topic classification. Therefore it is natural to raise the question: is it possible to take advantage of the domain knowledge instilled in pattern extractors during the fine-tuning process of a pre-trained language model?  This study investigates this research question in a financial domain in industrial settings.

Our work is also related to weak supervision \cite{DBLP:conf/acl/MekalaS20, DBLP:conf/naacl/KaramanolakisMZ21}. However our approach is different from the weak supervision framework in the following way: we don't employ regular expression to first generate weak label and then train the neural topic classification model with those weak labels. Instead, we input the regular expression generated features as an auxiliary input in parallel with the dense features learned from BERT during fine-tuning to help the domain adaptation process.

\section{Methodology} \label{sec:methodology}

\begin{figure*}[h!]
  \centering
  \includegraphics[width=0.98\textwidth]{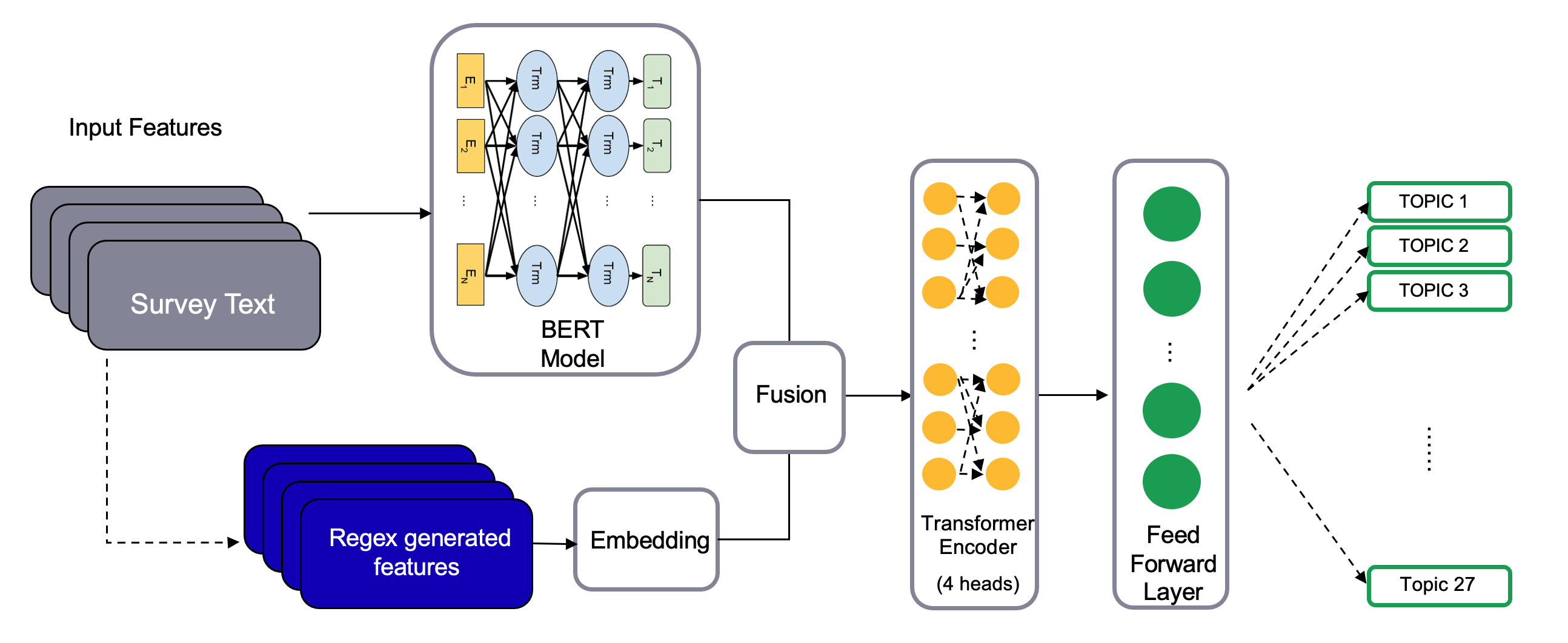}
  \caption{Model Structure Overview}
  \label{fig:Model}
\end{figure*}

Our methodology for fine tuning of a pre-trained BERT model is shown in Figure \ref{fig:Model}. First, we pass text surveys as sequences to a pre-trained BERT model and extract the CLS representation vectors for text sequences \cite{DBLP:journals/corr/abs-1903-10318}  \cite{DBLP:conf/naacl/DevlinCLT19} . Our CLS representation is actually a pooled representation that is further processed by linear layer and Tanh activation function \footnote[1]{\url{https://huggingface.co/transformers/v3.0.2/model\_doc/bert.html\#bertmodel}}. This is a common approach used in the literature for the extraction of contextual representation of a text sequence \cite{DBLP:journals/corr/abs-1910-05032}. Second, we use regular expressions (regex) based rules to assign each text survey a topic feature. We have 27 regular expressions, one for each topic. Each survey can have multiple topic features generated by regular expressions. We have a minimum of zero and a maximum of seven topic features for our surveys. Regular expressions are written by us with the help of finance and insurance domain experts for call centers. We show the description of these regular expressions in Table \ref{re-list} in appendix. A problem with regular expressions is that they would mistag if there is a slight variation of wordings in text. This means they don't generalize well on unseen data.  We therefore  convert these topic features created by regular expressions into embeddings and decided to fuse it with BERT-based representation of the text. Third, we concatenate BERT-based embeddings with embeddings for topic features. We call this step fusion. We concatenate all the topic embeddings assigned to text in previous step with padding vectors where necessary before concatenating them further with BERT embeddings for the corresponding text. Fourth, we pass this fused representation of vectors to a transformer encoder with four attention-heads to utilize the self-attention mechanism during fine-tuning \cite{DBLP:journals/corr/VaswaniSPUJGKP17}. Fifth, we pass this to a two layer feed forward neural network with ReLU activation on the first layer and 27 sigmoid units on the output layer. We have framed this problem as a multi-label classification problem as each text survey can have multiple topic labels. Finally, we fine tune this neural network architecture on a small dataset. During inference time, we use this model to predict 27 topics for a survey. If all predictions are below a certain threshold, determined based on human judgement using test set, we mark the topic as emerging topic (the 28th topic). Our experimental details are further shown in the next section.

\section{Experimental Setup}\label{sec:experiments}
In this section, we will present in detail our experimental setup, model configuration, evaluation results and discussions.

\subsection{Dataset}
In this study, we employ a private dataset collected through customer service feedback survey. Each sample in this dataset contains a text snippet which is the written feedback from a customer, and topic labels from a pre-defined set of 27 topics. The distribution of these topic labels for train and test dataset are listed in Table \ref{dataset distribution}. Those topics cover Agent Communication, Agent Service Quality, Audio Call Quality, Product Coverage, Claim Result, Ease-to-use of the Website, etc. The dataset has in total 541 samples, and the average length of the text is 47. We have manually labelled these 541 samples with the help of domain experts.

\subsection{Evaluation Metrics and Protocol}
In order to determine the best hyper-parameter settings, we monitor the cross entropy loss on the validation set during training. For the evaluation on testset, we employ weighted F1 Score \cite{DBLP:conf/fire/0001MMPDMP19} as our evaluation metrics, considering each of our targeting topics has an imbalanced distribution. The calculation of the weighted F1 score is presented in Equation \ref{eq:weighted_f1}. We first calculate the F1 score for each class $F_{1i}$. Second we take the weighted average of class-specific F1 scores $F_{1i}$ to get the weighted F1 score. The weight for each class $i$ is calculated based on the support of the class $i$, as shown in the following equation.

\begin{align}
    Weighted\_F_1 &= \sum_{i=1}^{K} w_{i} F_{1i}  \\
    w_i &= \frac{support_i}{\sum_{i=1}^{K}support_i}
    \label{eq:weighted_f1}
\end{align}

For the purpose of training, validation and evaluation of our proposed model, we split the whole dataset into training and testing subsets with a ratio of 7:3. To simulate as much as possible the real use case, the test instances in the test set was sampled according to the distribution of the real usage scenario. It contains $127$ test instances which could be easily classified by regular expressions, and another $35$ test instances which could not be classified by regular expressions. This distribution is very close to the real application data distribution that our business partner is dealing with in their daily work.

%For the training of the main model, we enhanced the data labeling quality and adopted four-fold cross validation protocol to split our dataset: We holdout 25\% data as test fold, and further split the remaining 75\% data into 3 fold and run cross-validation. Four different models were trained and we selected the best one for production purpose.

\subsection{Model Configurations}
We set the max sequence length of the BERT tokenizer to be 250, and keep the original BERT's hidden layer size of 768. We employ the AdamW optimizer \cite{DBLP:journals/corr/abs-2202-00089} and set the initial learning rate to be $2e-5$. During training, the batch size is set to 8. The number of the regular expression generated discrete features is 28, 27 topic features from regular expressions as shown in Table \ref{re-list}  and one is no-topic if none of the regex matches the input text. The number of self-attention heads in the feature fusion layer is set to 4. We set the maximum number of training epochs to be 30.

%We have 27 independent regex matching function to detect whether a text belong to one of the topic classes, one ``emerging class" and another zero class for padding purpose. Those discrete regex features are then embedded into dense regex class embedding vectors of dimension 28. We initialize the regex class embeddings randomly from a normal distribution $\mathcal{N}(0, 1)$ Those embedded regex class vectors are then reshaped into a long flattened vector to be concatenated with the BERT-learned representation.

%With the concatenated vectors, we applied an attention layer to it so that the model can learn the weight of regex generated features and text information. The number of heads were set at 4.
%Then we applied feed forward layers with layernorm, relu dropped out layer. And we feed the output into a linear layers and Sigmoid to shape the output into batch size by number of prediction classes.

We develop our model based on Pytorch\footnote[1]{\url{https://pytorch.org}} and Pytorch Lightning\footnote[2]{\url{https://www.pytorchlightning.ai}} packages. The training and testing was performed on Azure Machine Learning Platform \footnote[3]{\url{https://azure.microsoft.com/en-ca/services/machine-learning/}} with a cluster of 6 cores CPU (112 GB RAM, 336 GB disk) and one NVIDIA Tesla V100 GPU. 

% \begin{figure}[h!]
%   \centering
%   \includegraphics[width=0.95\textwidth]{images/model-config.png}
%   \caption{Model alternative configurations}
%   \label{fig:alternative configuration}
% \end{figure}

%compared with just regulation.

\subsection{Evaluation Results and Discussion}\label{sec:eval}

\begin{table*}[h!]
  \caption{Evaluation Results}
  \label{tab:results}
  \centering
  \scalebox{0.78}{
  \begin{tabular}{ccccccc}
    \toprule
    \cmidrule(r){1-2}
     Models	& Input Channel1 (text)	& Input Channel 2 (regex) & Fusion Layer & Precision & Recall &	Weighted  F1-score \\
    \midrule
    1  & N/A      &   REGEX Generated Features  & N/A    &     0.76  &  0.54  &  0.58       \\
    2  & BERT MODEL & N/A               &  SelfAttention &  0.44  &  0.79  &  0.53  \\
    \midrule
    3  & BERT MODEL & RegexEmbeddingBag &  SelfAttention &  0.44  &  0.79  &  0.55  \\
    4  & BERT MODEL & RegexEmbedding    &  Linear    &  0.49  &  0.81  &  0.59  \\
     %  & BERT MODEL(freeze para) & RegexEmbedding & Attention & 0.49 & 0.75 & 0.57 \\
    5  & BERT MODEL & RegexEmbedding    &  SelfAttention &  0.55  &  \textbf{0.83}  &  \textbf{0.64}  \\
    \bottomrule
  \end{tabular}}
%   \label{tab:results}
\end{table*}

\noindent\textbf{Evaluation Results}: The evaluation results are presented in Table \ref{tab:results}. Model 1 relies solely on regex expressions to classify topics and doesn't have any machine learning components. Model 2 relies solely on the BERT hidden representation to classify topics and employs a self-attention layer before the output layer. Those two models are our baselines to compare with. The lower part of Table \ref{tab:results} shows the evaluation results of our main model and its variants: Model 3 and Model 5 employ both the BERT hidden representation and the regex generated embedding features as input sources. They differ in the regex feature embedding transformations: The Model 3 employs a bagged embedding transformation which average-pools all the regex embedding vectors, whereas the Model 5 employs an ordinary embedding layer and concatenates all the regex embeddings. Both of them employ a transformer-based self-attention layer to fuse the BERT-learned representation and the regex-generated embeddings. To study the effect on the choice of the fusion layer, we also have Model 4 to be compared with Model 5, both of which share the same input sources but differ in the fusion layer. 

From Table \ref{tab:results}, we first observe that under the best configuration, our Model 5 outperforms both baselines (Model 1 and Model 2) on weighted F1 score by large margins. This validates our hypothesis that employing regex-generated features is helpful for the fine-tuning process in financial domain. We observe relative improvements \footnotemark[4] of $10.34\%$ and $20.75\%$ on weighted F1-score of Model 5 over Model 1 and Model 2 respectively. This comparison confirms that combining both BERT-learned representations and regex embeddings are more advantageous than using either BERT representation or Regex features alone.

\noindent\textbf{Discussions}: The Model 1 which employs solely regex to classify topics performs poorly. This is possibly because regular expressions are only good at catching exact matched patterns but lacks the power of generalization. If the wording in user's feedback slightly changes from our predefined regex rules, this model may fail to classify this sample correctly. The Model 2 which employs solely BERT-learned representations performs poorly as well. This is possibly due to the fact that BERT-learned representations are very good at capturing semantically related concepts but lacks the capability to identify exact match. By combining both the semantic and regex features, our proposed model could handle both the cases of exact matched rules and semantically related concepts. 

We also study the effect of choosing different fusion layers on classification performance. In both Model 4 and Model 5, we combine BERT-learned semantic representations and regex-generated features, but choose different fusion layers: The Model 4 employs a simple linear transformation layer whereas the Model 5 employs a self-attention layer. In Table \ref{tab:results}, on a relative improvement scale\footnote[4]{relative improvement is calculated as (score\_y - score\_x) / score\_x}, we observe that Model 5 outperformed Model 4 by $12.24\%$ on precision and $8.47\%$ on weighted F1 score, and $2.47\%$ on recall score. This comparison demonstrates that employing self-attention layer as fusion layer is more effective than a simple linear layer, since self-attention heads could attend more to the important features given the current context.

\noindent\textbf{Case Study}: 
\begin{table}[h!]
  \caption{Case Study Example}
  \label{tab:examples}
  \centering
  \scalebox{0.70}{
  
  \begin{tabular}{p{0.3\linewidth} | p{0.2\linewidth} | p{0.2\linewidth} | p{0.2\linewidth} | p{0.2\linewidth}}
  
    \toprule
    \cmidrule(r){1-2}
     Survey Text	& Regex only  (Model 1) & BERT only  (Model 2) & Regex +BERT (Model 5) & Ground Truth Labels\\
    \midrule
     I would like to have my coverage details emailed to me which still was not done. There was no proof for the explanation provided by your agent.  &	`Emerging Topic'	 & 
     `Call Transfer', `Called Multiple Times', `Call Lack of Follow-ups', `Unable to Resolve Issue', `Claim Process'&
    `Call Lack of Follow-ups' &	`Call Lack of Follow-ups'     \\
    \bottomrule
  \end{tabular}}
\end{table}
To better showcase the advantage of our combined model over regex-only or BERT-only models during fine-tuning, we present a test sample together with the inference results of Model 1, Model 2, Model 5 and the ground truth labels in Table \ref{tab:examples}. In this example the customer complains about the lack of follow-ups after an insurance coverage inquiry. We can observe that the Regex-only Model 1 could not classify it with any of the pre-defined 27 topics, and outputs an `Emerging Topic' as result. The BERT-only Model 2 over-generalizes the semantics of the text and outputs many irrelevant topics other than the correct one. When we combine regex-generated features and BERT-learned representations during fine-tuning, our combination Model 5 outputs the correct topic label without any over-generalization. This example demonstrates that combining regex-generated features does help the fine-tuning process of a pre-trained language model in our specialized domain. The overall performance comparison of the regex-only, BERT-only and combined model is presented in Table \ref{tab:results}.

\noindent\textbf{Production Deployment and User Feedback}:
To empower users to benefit from the modeling result quickly and effectively and unearth actionable insights, we develop an application which integrates the following processes in an end-to-end pipeline: 1) Survey texts are automatically collected through third party API and are cleaned up by the prepossessing script. 2) The classification model described in Sec \ref{sec:methodology} takes as input the prepossessed survey  and assign topics to the text. 3) An automated connection to Elasticsearch \footnote[1]{https://www.elastic.co/} is established to push the predicted results to Elastic Database. 4) Finally the prediction results are presented to users through Kibana dashboard \footnote[2]{https://www.elastic.co/kibana/kibana-dashboard} for users to interact with the survey and predicted topics.
%survey text data  through third party API, process the survey data by cleaning up, perform topic classification and assign the topic. An connection to Elasticsearch is established to push the generated results to Elastic database. The data is then pushed to Elastic Kibana dashboard for user to interact with the survey data and generated topics.

In order to evaluate the performance of our proposed regex-augmented BERT model in the real production scenario, we design a questionnaire consisting of 6 questions, and ask the users to rate based on a scale of $1$ to $10$ for each question. The questionnaire is presented in Table \ref{tab:questionnaire} in appendix. We interviewed all end users with this questionnaire and asked them to rate the model performance improvement of the regex-augmented BERT model over the regex-only model in terms of topic coverage (recall), prediction precision, and time savings. Our proposed regex-augmented BERT model, when put into production and evaluated by users' subjective feedback, improves in average $40.00\%$ on user ratings for topic coverage , $58.33\%$ on user ratings for prediction precision  and $26.67\%$ on time savings over the previous regex-only model according to user ratings. These significant improvements on user evaluations demonstrate again the effectiveness of our proposed model, and showcase that proposed model is indeed powerful in a real-life production environment.

%by comparing the Regex only model which users have been using it since year 2019 to 2021 with the Regex+BERT model which we implemented in year 2022. Our users do see a significant improvement in both model coverage and classification accuracy.

\section{Conclusion and Future Work}

In this paper, we demonstrated a fine tuning method of a pre-trained language model in finance and insurance domain using features generated with regular expression. Our experiment results demonstrated the effectiveness of involving regular expression to boost the fine-tuning performance. We also learnt that the use of attention network, like encoder transformer, can help boost the performance of fine tuning compared to linear layers. We hope that this paper could offer insights to community into the fine-tuning process of a pre-trained language model with domain-specific data.

The limitation of our study is that we only used a small dataset to evaluate our results in one specialized domain. Our results require validation on different domains. In future, we plan to extend this work on other domains, such as marketing and advertisement, and determine the effective methods to fine tune a model. We also need to explore the use of other feature engineering methods that can be combined in the fine tuning phase effectively.

\bibliographystyle{plainnat}
\bibliography{refs}

\begin{thebibliography}{23}
\providecommand{\natexlab}[1]{#1}
\providecommand{\url}[1]{\texttt{#1}}
\expandafter\ifx\csname urlstyle\endcsname\relax
  \providecommand{\doi}[1]{doi: #1}\else
  \providecommand{\doi}{doi: \begingroup \urlstyle{rm}\Url}\fi

\bibitem[AlBatayha(2021)]{albatayha2021multi}
Duha AlBatayha.
\newblock Multi-topic labelling classification based on lstm.
\newblock In \emph{2021 12th International Conference on Information and
  Communication Systems (ICICS)}, pages 471--474. IEEE, 2021.

\bibitem[Chauhan and Shah(2022)]{DBLP:journals/csur/ChauhanS22}
Uttam Chauhan and Apurva Shah.
\newblock Topic modeling using latent dirichlet allocation: {A} survey.
\newblock \emph{{ACM} Comput. Surv.}, 54\penalty0 (7):\penalty0 145:1--145:35,
  2022.

\bibitem[Chen et~al.(2019)Chen, Ma, Harimoto, Bao, Su, and
  Sun]{DBLP:journals/corr/abs-1910-05032}
Deli Chen, Shuming Ma, Keiko Harimoto, Ruihan Bao, Qi~Su, and Xu~Sun.
\newblock Group, extract and aggregate: Summarizing a large amount of finance
  news for forex movement prediction.
\newblock \emph{CoRR}, abs/1910.05032, 2019.
\newblock URL \url{http://arxiv.org/abs/1910.05032}.

\bibitem[Devlin et~al.(2019)Devlin, Chang, Lee, and
  Toutanova]{DBLP:conf/naacl/DevlinCLT19}
Jacob Devlin, Ming{-}Wei Chang, Kenton Lee, and Kristina Toutanova.
\newblock {BERT:} pre-training of deep bidirectional transformers for language
  understanding.
\newblock In Jill Burstein, Christy Doran, and Thamar Solorio, editors,
  \emph{Proceedings of the 2019 Conference of the North American Chapter of the
  Association for Computational Linguistics: Human Language Technologies,
  {NAACL-HLT} 2019, Minneapolis, MN, USA, June 2-7, 2019, Volume 1 (Long and
  Short Papers)}, pages 4171--4186. Association for Computational Linguistics,
  2019.

\bibitem[Ghourabi(2021)]{ghourabi2021bert}
Abdallah Ghourabi.
\newblock A bert-based system for multi-topic labeling of arabic content.
\newblock In \emph{2021 12th International Conference on Information and
  Communication Systems (ICICS)}, pages 486--489. IEEE, 2021.

\bibitem[Grootendorst(2022)]{DBLP:journals/corr/abs-2203-05794}
Maarten Grootendorst.
\newblock Bertopic: Neural topic modeling with a class-based {TF-IDF}
  procedure.
\newblock \emph{CoRR}, abs/2203.05794, 2022.

\bibitem[Karamanolakis et~al.(2021)Karamanolakis, Mukherjee, Zheng, and
  Awadallah]{DBLP:conf/naacl/KaramanolakisMZ21}
Giannis Karamanolakis, Subhabrata Mukherjee, Guoqing Zheng, and Ahmed~Hassan
  Awadallah.
\newblock Self-training with weak supervision.
\newblock In Kristina Toutanova, Anna Rumshisky, Luke Zettlemoyer, Dilek
  Hakkani{-}T{\"{u}}r, Iz~Beltagy, Steven Bethard, Ryan Cotterell, Tanmoy
  Chakraborty, and Yichao Zhou, editors, \emph{Proceedings of the 2021
  Conference of the North American Chapter of the Association for Computational
  Linguistics: Human Language Technologies, {NAACL-HLT} 2021, Online, June
  6-11, 2021}, pages 845--863. Association for Computational Linguistics, 2021.

\bibitem[Lewis et~al.(2020)Lewis, Liu, Goyal, Ghazvininejad, Mohamed, Levy,
  Stoyanov, and Zettlemoyer]{DBLP:conf/acl/LewisLGGMLSZ20}
Mike Lewis, Yinhan Liu, Naman Goyal, Marjan Ghazvininejad, Abdelrahman Mohamed,
  Omer Levy, Veselin Stoyanov, and Luke Zettlemoyer.
\newblock {BART:} denoising sequence-to-sequence pre-training for natural
  language generation, translation, and comprehension.
\newblock In Dan Jurafsky, Joyce Chai, Natalie Schluter, and Joel~R. Tetreault,
  editors, \emph{Proceedings of the 58th Annual Meeting of the Association for
  Computational Linguistics, {ACL} 2020, Online, July 5-10, 2020}, pages
  7871--7880. Association for Computational Linguistics, 2020.

\bibitem[Liu et~al.(2020)Liu, Xia, Li, Yan, and Liu]{liu2020bert}
Jingang Liu, Chunhe Xia, Xiaojian Li, Haihua Yan, and Tengteng Liu.
\newblock A bert-based ensemble model for chinese news topic prediction.
\newblock In \emph{Proceedings of the 2020 2nd International Conference on Big
  Data Engineering}, pages 18--23, 2020.

\bibitem[Liu(2019)]{DBLP:journals/corr/abs-1903-10318}
Yang Liu.
\newblock Fine-tune {BERT} for extractive summarization.
\newblock \emph{CoRR}, abs/1903.10318, 2019.
\newblock URL \url{http://arxiv.org/abs/1903.10318}.

\bibitem[Liu et~al.(2019)Liu, Ott, Goyal, Du, Joshi, Chen, Levy, Lewis,
  Zettlemoyer, and Stoyanov]{DBLP:journals/corr/abs-1907-11692}
Yinhan Liu, Myle Ott, Naman Goyal, Jingfei Du, Mandar Joshi, Danqi Chen, Omer
  Levy, Mike Lewis, Luke Zettlemoyer, and Veselin Stoyanov.
\newblock Roberta: {A} robustly optimized {BERT} pretraining approach.
\newblock \emph{CoRR}, abs/1907.11692, 2019.
\newblock URL \url{http://arxiv.org/abs/1907.11692}.

\bibitem[Mandl et~al.(2019)Mandl, Modha, Majumder, Patel, Dave, Mandalia, and
  Patel]{DBLP:conf/fire/0001MMPDMP19}
Thomas Mandl, Sandip Modha, Prasenjit Majumder, Daksh Patel, Mohana Dave,
  Chintak Mandalia, and Aditya Patel.
\newblock Overview of the {HASOC} track at {FIRE} 2019: Hate speech and
  offensive content identification in indo-european languages.
\newblock In Prasenjit Majumder, Mandar Mitra, Surupendu Gangopadhyay, and
  Parth Mehta, editors, \emph{{FIRE} '19: Forum for Information Retrieval
  Evaluation, Kolkata, India, December, 2019}, pages 14--17. {ACM}, 2019.

\bibitem[Mekala and Shang(2020)]{DBLP:conf/acl/MekalaS20}
Dheeraj Mekala and Jingbo Shang.
\newblock Contextualized weak supervision for text classification.
\newblock In Dan Jurafsky, Joyce Chai, Natalie Schluter, and Joel~R. Tetreault,
  editors, \emph{Proceedings of the 58th Annual Meeting of the Association for
  Computational Linguistics, {ACL} 2020, Online, July 5-10, 2020}, pages
  323--333. Association for Computational Linguistics, 2020.

\bibitem[Mozafari et~al.(2019)Mozafari, Farahbakhsh, and
  Crespi]{DBLP:conf/complexnetworks/MozafariFC19}
Marzieh Mozafari, Reza Farahbakhsh, and No{\"{e}}l Crespi.
\newblock A bert-based transfer learning approach for hate speech detection in
  online social media.
\newblock In Hocine Cherifi, Sabrina Gaito, Jos{\'{e}}~Fernendo Mendes, Esteban
  Moro, and Luis~Mateus Rocha, editors, \emph{Complex Networks and Their
  Applications {VIII} - Volume 1 Proceedings of the Eighth International
  Conference on Complex Networks and Their Applications {COMPLEX} {NETWORKS}
  2019, Lisbon, Portugal, December 10-12, 2019}, volume 881 of \emph{Studies in
  Computational Intelligence}, pages 928--940. Springer, 2019.

\bibitem[Peng et~al.(2018)Peng, Li, He, Liu, Bao, Wang, Song, and
  Yang]{DBLP:conf/www/PengLHLBWS018}
Hao Peng, Jianxin Li, Yu~He, Yaopeng Liu, Mengjiao Bao, Lihong Wang, Yangqiu
  Song, and Qiang Yang.
\newblock Large-scale hierarchical text classification with recursively
  regularized deep graph-cnn.
\newblock In Pierre{-}Antoine Champin, Fabien Gandon, Mounia Lalmas, and
  Panagiotis~G. Ipeirotis, editors, \emph{Proceedings of the 2018 World Wide
  Web Conference on World Wide Web, {WWW} 2018, Lyon, France, April 23-27,
  2018}, pages 1063--1072. {ACM}, 2018.

\bibitem[Rubin et~al.(2012)Rubin, Chambers, Smyth, and
  Steyvers]{DBLP:journals/ml/RubinCSS12}
Timothy~N. Rubin, America Chambers, Padhraic Smyth, and Mark Steyvers.
\newblock Statistical topic models for multi-label document classification.
\newblock \emph{Mach. Learn.}, 88\penalty0 (1-2):\penalty0 157--208, 2012.

\bibitem[Sun et~al.(2019)Sun, Qiu, Xu, and Huang]{DBLP:conf/cncl/SunQXH19}
Chi Sun, Xipeng Qiu, Yige Xu, and Xuanjing Huang.
\newblock How to fine-tune {BERT} for text classification?
\newblock In Maosong Sun, Xuanjing Huang, Heng Ji, Zhiyuan Liu, and Yang Liu,
  editors, \emph{Chinese Computational Linguistics - 18th China National
  Conference, {CCL} 2019, Kunming, China, October 18-20, 2019, Proceedings},
  volume 11856 of \emph{Lecture Notes in Computer Science}, pages 194--206.
  Springer, 2019.

\bibitem[Vaswani et~al.(2017)Vaswani, Shazeer, Parmar, Uszkoreit, Jones, Gomez,
  Kaiser, and Polosukhin]{DBLP:journals/corr/VaswaniSPUJGKP17}
Ashish Vaswani, Noam Shazeer, Niki Parmar, Jakob Uszkoreit, Llion Jones,
  Aidan~N. Gomez, Lukasz Kaiser, and Illia Polosukhin.
\newblock Attention is all you need.
\newblock \emph{CoRR}, abs/1706.03762, 2017.
\newblock URL \url{http://arxiv.org/abs/1706.03762}.

\bibitem[Wang and Manning(2012)]{DBLP:conf/acl/WangM12}
Sida Wang and Christopher~D. Manning.
\newblock Baselines and bigrams: Simple, good sentiment and topic
  classification.
\newblock In \emph{The 50th Annual Meeting of the Association for Computational
  Linguistics, Proceedings of the Conference, July 8-14, 2012, Jeju Island,
  Korea - Volume 2: Short Papers}, pages 90--94. The Association for Computer
  Linguistics, 2012.

\bibitem[Whang et~al.(2020)Whang, Lee, Lee, Yang, Oh, and
  Lim]{DBLP:conf/interspeech/WhangLLYOL20}
Taesun Whang, Dongyub Lee, Chanhee Lee, Kisu Yang, Dongsuk Oh, and Heuiseok
  Lim.
\newblock An effective domain adaptive post-training method for {BERT} in
  response selection.
\newblock In Helen Meng, Bo~Xu, and Thomas~Fang Zheng, editors,
  \emph{Interspeech 2020, 21st Annual Conference of the International Speech
  Communication Association, Virtual Event, Shanghai, China, 25-29 October
  2020}, pages 1585--1589. {ISCA}, 2020.

\bibitem[Yang et~al.(2022)Yang, Zhang, Qin, Li, Wang, Liu, Wang, Xie, and
  Zhang]{DBLP:journals/corr/abs-2211-08073}
Linyi Yang, Shuibai Zhang, Libo Qin, Yafu Li, Yidong Wang, Hanmeng Liu, Jindong
  Wang, Xing Xie, and Yue Zhang.
\newblock {GLUE-X:} evaluating natural language understanding models from an
  out-of-distribution generalization perspective.
\newblock \emph{CoRR}, abs/2211.08073, 2022.

\bibitem[Yang et~al.(2019)Yang, Dai, Yang, Carbonell, Salakhutdinov, and
  Le]{DBLP:conf/nips/YangDYCSL19}
Zhilin Yang, Zihang Dai, Yiming Yang, Jaime~G. Carbonell, Ruslan Salakhutdinov,
  and Quoc~V. Le.
\newblock Xlnet: Generalized autoregressive pretraining for language
  understanding.
\newblock In Hanna~M. Wallach, Hugo Larochelle, Alina Beygelzimer, Florence
  d'Alch{\'{e}}{-}Buc, Emily~B. Fox, and Roman Garnett, editors, \emph{Advances
  in Neural Information Processing Systems 32: Annual Conference on Neural
  Information Processing Systems 2019, NeurIPS 2019, 8-14 December 2019,
  Vancouver, BC, Canada}, pages 5754--5764, 2019.

\bibitem[Zhuang et~al.(2022)Zhuang, Liu, Cutkosky, and
  Orabona]{DBLP:journals/corr/abs-2202-00089}
Zhenxun Zhuang, Mingrui Liu, Ashok Cutkosky, and Francesco Orabona.
\newblock Understanding adamw through proximal methods and scale-freeness.
\newblock \emph{CoRR}, abs/2202.00089, 2022.
\newblock URL \url{https://arxiv.org/abs/2202.00089}.

\end{thebibliography}

\appendix
\section{Appendix}

\begin{table*}[htbp]
\caption{User Feedback Survey Questionnaire}
  \centering
  \scalebox{0.79}{
    \begin{tabular}{ccp{29.25em}}
    \toprule
    \multicolumn{1}{c}{Categories} & \multicolumn{1}{l}{No.} & \multicolumn{1}{l}{Questions} \\
    \midrule
    \multicolumn{1}{c}{\multirow{3}[6]{*}{NPS360 old Regex Model (2019-2021)}} & 1     & How satisfied are you with the generated topics’ coverage (recall) of NPS360 application with the older regex model? \\
\cmidrule{2-3}          & 2     & How satisfied are you with the generated topics’ precision of NPS360 application with the older regex model? \\
\cmidrule{2-3}          & 3     & How do you rate the time \& effort savings through the usage of the NPS360 application’s older regex model?  \\
    \midrule
    \multicolumn{1}{c}{\multirow{3}[6]{*}{NPS360 New Model (2022)}} & 4     & How satisfied are you with the generated topics’ coverage (recall) of NPS360 application with the new model? \\
\cmidrule{2-3}          & 5     & How satisfied are you with the generated topics’ precision of NPS360 application with the new model? \\
\cmidrule{2-3}          & 6     & How do you rate the time \& effort savings through the usage of the NPS360 application’s new model?  \\
    \bottomrule
    \end{tabular}}%
  \label{tab:questionnaire}%
\end{table*}%

\begin{table*}[h!]
  \caption{Description of the regular expression functions}
  \label{re-list}
  \centering
  \scalebox{0.66}{
  \begin{tabular}{p{0.03\linewidth} | p{0.45\linewidth} | p{1.0\linewidth}}
    \toprule
    \cmidrule(r){1-2}
     No. & Regex Generated Feature & Description of Regex Expression Function Used \\
    \midrule
    1	&	Agent Communication	&	If customer cannot understand the agent, language difficulty/barrier, if the agent cannot speak fluently	\\
    2	&	Agent Knowledge and Resolution Speed	&	When the agent can/cannot provide accurate information/answers on what the customer asks, can/cannot provide info quickly	\\
    3	&	Offshore Agents	& Customers feedback about offshore service agents \\
    4	&	Agent Service Attitude	&	About the agents behaviour during the call	\\
    5	&	Agent Service Quality	&	Overal service quality customer receives	\\
    6	&	Application Authentication	&	Troubles with logging into the app, password reset and similar issues	\\
    7	&	Application Interface/Navigation	&	Being or not being able to easily navigate the app and having a good/bad interface or experience while using the app	\\
    8	&	Audio Quality/Connection	&	poor connections and customer unable to hear	\\
    9	&	Call Transfer	&	call was transferred into other departments or multiple times	\\
    10	&	Called Multiple Times & customer called multiple times	\\
    11	&	IVR	&	phone prompt system or menu	\\
    12	&	Call Lack of follow-up	&	customer service didn't call back and follow up with customer	\\
    13	&	Language Barrier	&	communication issue with the service agent	\\
    14	&	Unable to resolve issue	&	call was not able to resolve the issue	\\
    15	&	Claims Online Submission	&	if customer has issue on how to submit claims online	\\
    16	&	Claims Process	&	claim processing time is too long	\\
    17	&	Claims Result	&	claim decisions are declined or incorrect	\\
    18	&	Claims Status/Decision	&	claim status or decisions	\\
    19	&	Claims Payment/Reimbursement	&	claim payment and reimbursement status	\\
    20	&	Product Coverage/Policy	& customer checking coverage details or has concern on coverage amount	\\
    21	&	Portal Authentication	&	Similar to app authentication but with respect to website/portal/site	\\
    22	&	Portal Computer Literacy	&	Having knowledge on how to use computers, being or not being tech savvy	\\
    23	&	Portal Ease of use	&	If the portal is easy to use for customer	\\
    24	&	Portal Information	&	The website not having enough information on what the customer was searching for	\\
    25	&	Portal Interface/Navigation	&	Trouble with the website interface itself if it is not user friendly or is user friendly	\\
    26	&	Portal Password	&	Specific issues related to password, customer not being able to reset them or entering the password doesn’t work on the websites	\\
    27	&	Portal Stability/Loading Time	&	Website crashes, loading time, down time or some kind of operational issues with the website	\\
    %\bottomrule
  \end{tabular}}
\end{table*}

\begin{table*}[!h]
  \caption{Topic Labels Distribution}
  \label{dataset distribution}
  \centering
  \scalebox{0.79}{
  \begin{tabular}{cccc}
    \toprule
     No.	&	Topic Labels & Train dataset label distribution	& Test dataset label distribution \\
    \midrule
1	&	Agent Communication						& 28 (3.5\%)	& 12 (3.6\%) \\
2	&	Agent Knowledge and Resolution Speed	& 47 (5.9\%)	& 34 (10.1\%) \\
3	&	Offshore Agents							& 7	 (0.9\%)	& 2	 (0.6\%) \\
4	&	Agent Service Attitude					& 65 (8.2\%)	& 33 (9.8\%) \\
5	&	Agent Service Quality					& 53 (6.7\%)	& 41 (12.2\%) \\
6	&	Application Authentication				& 25 (3.2\%)	& 7	 (2.1\%) \\
7	&	Application Interface/Navigation		& 17 (2.1\%)	& 4	 (1.2\%) \\
8	&	Audio Quality/Connection				& 27 (3.4\%)	& 13 (3.9\%) \\
9	&	Call Transfer							& 26 (3.3\%)	& 7	 (2.1\%) \\
10	&	Called Multiple Times					& 31 (3.9\%)	& 12 (3.6\%) \\
11	&	IVR										& 10 (1.3\%)	& 9	 (2.7\%) \\
12	&	Call Lack of follow-up					& 24 (3.0\%)	& 20 (6.0\%) \\
13	&	Language Barrier						& 22 (2.8\%)	& 3	 (0.9\%) \\
14	&	Unable to resolve issue					& 29 (3.7\%)	& 13 (3.9\%) \\
15	&	Claims Online Submission				& 22 (2.8\%)	& 11 (3.3\%) \\
16	&	Claims Process							& 29 (3.7\%)	& 11 (3.3\%) \\
17	&	Claims Result							& 34 (4.3\%)	& 7	 (2.1\%) \\
18	&	Claims Status/Decision					& 27 (3.4\%)	& 3	 (0.9\%) \\
19	&	Claims Payment/Reimbursement			& 27 (3.4\%)	& 5	 (1.5\%) \\
20	&	Product Coverage/Policy					& 20 (2.5\%)	& 7	 (2.1\%) \\
21	&	Portal Authentication					& 47 (5.9\%)	& 13 (3.9\%) \\
22	&	Portal Computer Literacy				& 2	 (0.3\%)	& 0	 (0.0\%) \\
23	&	Portal Ease of use						& 41 (5.2\%)	& 21 (6.3\%) \\
24	&	Portal Information						& 27 (3.4\%)	& 11 (3.3\%) \\
25	&	Portal Interface/Navigation				& 38 (4.8\%)	& 11 (3.3\%) \\
26	&	Portal Password							& 39 (4.9\%)	& 14 (4.2\%) \\
27	&	Portal Stability/Loading Times			& 29 (3.7\%)	& 12 (3.6\%) \\
    
    \bottomrule
  \end{tabular}}
\end{table*}

\clearpage
\onecolumn

\end{document}